\documentclass[10pt,leqno]{amsart}
\usepackage{graphicx}
\usepackage{indentfirst,csquotes}
\usepackage{pgfplots}
\usepackage{tcolorbox}
\usepackage{graphicx}
\usepackage{indentfirst,csquotes}
\usepackage{amssymb,amsthm,amsmath}
\usepackage{xcolor,paralist,hyperref,fancyhdr,etoolbox} 
\usepackage{braket}
\usepackage{subcaption} 
\usepackage{booktabs} 
\usepackage{tikz}
\usepackage{standalone}
\usepackage{pgfplots}
\usepackage{lmodern}
\usepackage{import}
\usepackage{algorithm}
\usepackage{algpseudocode}
\usepackage{graphicx}
\usepackage{amssymb,amsmath,mathtools}
\usepackage{hyperref}
\usepackage[T1]{fontenc}
\usepackage{braket}
\usepackage{booktabs}
\usepackage{tcolorbox}   
\usepackage{tikz}        
\usepackage{listings}    
\usepackage{xcolor}      
\usepackage{tikz}
\usetikzlibrary{positioning, arrows.meta, shapes.geometric}

\tikzstyle{block} = [rectangle, draw, text centered, rounded corners, minimum height=2em]
\tikzstyle{line} = [draw, -stealth, thick]
\tikzstyle{cloud} = [ellipse, draw, text centered, minimum height=2em, thick]

\usetikzlibrary{automata, positioning}
\usetikzlibrary{shapes.geometric, arrows, positioning}

\everymath=\expandafter{\the\everymath\displaystyle}

\tikzstyle{block} = [rectangle, draw, text centered, rounded corners, minimum height=2em]
\tikzstyle{line} = [draw, -stealth, thick]
\tikzstyle{cloud} = [ellipse, draw, text centered, minimum height=2em, thick]
\tikzstyle{dashedcloud} = [ellipse, draw, dashed, text centered, minimum height=2em, thick]
\usetikzlibrary{positioning, arrows.meta, shapes.geometric, decorations.pathreplacing}

\tikzstyle{startstop} = [rectangle, rounded corners, minimum width=1.5cm, minimum height=0.5cm,text centered, draw=black, fill=red!30]
\tikzstyle{io} = [trapezium, trapezium left angle=70, trapezium right angle=110, minimum width=1cm, minimum height=0.5cm, text centered, draw=black, fill=blue!30]

\tikzstyle{process} = [rectangle, minimum width=3cm, minimum height=0.5cm, text centered, draw=black, fill=orange!30]
\tikzstyle{decision} = [diamond, minimum width=0.5cm, minimum height=0.1cm, text centered, draw=black, fill=green!30]
\tikzstyle{process2} = [rectangle, minimum width=1cm, minimum height=0.5cm, text centered, draw=black, fill=orange!30]
\tikzstyle{arrow} = [thick,->,>=stealth]
\usetikzlibrary{shapes,shadows,arrows,positioning,angles,quotes}
\tikzset{My Arrow Style/.style={single arrow, fill=black!15, anchor=base, align=center,text width=2.3cm}}
\tikzstyle{arrow} = [thick,->,>=stealth]
\usetikzlibrary{calc,trees,positioning,arrows,fit,shapes,calc}

\newtheorem{theorem}{Theorem}
\newtheorem{definition}[theorem]{Definition}

\hypersetup{
  colorlinks=true,
  linkcolor=blue,    
  citecolor=blue,    
  filecolor=blue,    
  urlcolor=blue      
}

\begin{document}

\title{Formally Verifying and Explaining Sepsis Treatment Policies with COOL-MC}
\author[Dennis Gross]{Dennis Gross}
\thanks{Email:\quad\texttt{dennis@artigo.ai}  \quad\textbar\quad COOL-MC:\quad \url{https://github.com/LAVA-LAB/COOL-MC}}
\maketitle


\begin{abstract}
Safe and interpretable sequential decision-making is critical in healthcare, yet \emph{reinforcement learning (RL)} policies for sepsis treatment optimization remain opaque and difficult to verify. Standard probabilistic model checkers operate on the full state space, which becomes infeasible for larger MDPs, and cannot explain why a learned policy makes particular decisions. \emph{COOL-MC} wraps the model checker \emph{Storm} but adds three key capabilities: it constructs only the reachable state space induced by a trained policy, yielding a smaller discrete-time Markov chain amenable to verification even when full-MDP analysis is intractable; it automatically labels states with clinically meaningful atomic propositions; and it integrates explainability methods with \emph{probabilistic computation tree logic (PCTL)} queries to reveal which features drive decisions across treatment trajectories. We demonstrate \emph{COOL-MC's} capabilities on the ICU-Sepsis MDP, a benchmark derived from approximately 17,000 sepsis patient records, which serves as a case study for applying \emph{COOL-MC} to the formal analysis of sepsis treatment policies. Our analysis establishes hard bounds via full MDP verification, trains a safe RL policy that achieves optimal survival probability, and analyzes its behavior via PCTL verification and explainability on the induced DTMC. This reveals, for instance, that our trained policy relies predominantly on prior dosing history rather than the patient's evolving condition, a weakness that is invisible to standard evaluation but is exposed by \emph{COOL-MC}'s integration of formal verification and explainability. Our results illustrate how \emph{COOL-MC} could serve as a tool for clinicians to investigate and debug sepsis treatment policies before deployment.
\end{abstract}


\section{Introduction}
\emph{Safe and explainable sequential decision-making} is essential in healthcare~\cite{abdellatif2023reinforcement,yu2021reinforcement}, where treatment strategies must be trustworthy before deployment~\cite{miedema2026towards}.
\emph{Sepsis} is the third leading cause of death worldwide and the main cause of mortality in hospitals~\cite{komorowski2018artificial}. It occurs when the body's response to an infection spirals out of control, causing widespread inflammation that can lead to tissue damage, organ failure, and death~\cite{o2007sepsis}. Treating sepsis requires sequential clinical decisions tailored to each patient's evolving condition~\cite{choudhary2024icu}, which can be modeled as a \emph{Markov decision process (MDP)}~\cite{komorowski2018artificial,yu2023towards,choudhary2024icu} where, at each time step, the patient's clinical state is observed, a treatment action is selected, and the patient transitions to a new state.

Based on this MDP formulation, \emph{probabilistic model checking}~\cite{baier2008principles,baier2018model,DBLP:conf/birthday/KwiatkowskaN025} can verify properties expressed in \emph{probabilistic computation tree logic (PCTL)}~\cite{hansson1994logic}, a temporal logic for reasoning about the probability of event sequences over time. For instance, one can query the maximum survival probability under any policy, or analyze whether avoiding a particular treatment action affects patient outcomes.

However, for larger and more complex models, constructing and model-checking the full MDP becomes computationally infeasible due to the state-space explosion problem~\cite{DBLP:conf/allerton/KwiatkowskaNP10,DBLP:conf/setta/GrossJJP22,azeem20251}.

\emph{Reinforcement learning (RL)} offers a scalable alternative, where an agent learns a near-optimal policy concerning an objective by taking actions and receiving feedback through rewards and state observations from the MDP environment~\cite{sutton2018reinforcement,DBLP:conf/setta/GrossJJP22}.

Unfortunately, trained policies may exhibit \emph{unsafe behavior}~\cite{DBLP:conf/setta/GrossJJP22}, such as recommending wrong treatments~\cite{park2024consistent}, since rewards do not capture complex safety requirements~\cite{DBLP:journals/aamas/VamplewSKRRRHHM22}.
To ensure that learned policies respect safety requirements, \emph{safe RL} methods such as shielding~\cite{DBLP:conf/aaai/AlshiekhBEKNT18,DBLP:journals/cacm/KonighoferBJJP25} can be employed, which restrict the agent's choices during learning to prevent unsafe behavior.

However, when combined with \emph{deep neural networks}~\cite{DBLP:journals/nature/MnihKSRVBGRFOPB15}, the resulting policies remain opaque and difficult to explain~\cite{DBLP:conf/esann/GrossS24}, which is particularly problematic in safety-critical domains like healthcare~\cite{komorowski2018artificial}, where clinicians need to understand and trust a trained RL policy before acting on its recommendations.

\emph{COOL-MC} addresses this by combining safe RL with formal verification and explainability~\cite{DBLP:conf/setta/GrossJJP22,gross2025pctl,DBLP:conf/esann/GrossS24,gross2024enhancing}: after learning a policy via safe RL with shielding, it constructs only the reachable state space induced by that policy, yielding a \emph{discrete-time Markov chain (DTMC)} amenable to probabilistic model checking and explainable RL methods~\cite{DBLP:conf/setta/GrossJJP22}\footnote{Safe RL support was added to \emph{COOL-MC} after its original publication.}.
On the verification side, PCTL properties can be checked on the induced DTMC to quantify survival probabilities and characterize treatment trajectories.
On the explainability side, feature pruning removes individual state feature inputs and measures the resulting change in the policy's behavior~\cite{DBLP:conf/esann/GrossS24}, revealing which patient characteristics drive treatment decisions.
Additionally, state labeling via explainability methods such as feature-importance permutation ranking~\cite{breiman2001random} enables identification of which state features most strongly influence the policy's decisions across different treatment trajectories~\cite{gross2025pctl}.
Together, these approaches provide both formal guarantees about the policy's behavior and explanations of its sequential decision-making.

To date, however, COOL-MC's capabilities have not been applied to the verification and explanation of sepsis treatment policies.

In this paper, we demonstrate that \emph{COOL-MC} can be applied to a real-world healthcare setting using the ICU-Sepsis MDP provided by Choudhary et al.~\cite{choudhary2024icu} as a case study.
This MDP models personalized sepsis treatment in the \emph{intensive care unit (ICU)} as an MDP constructed from approximately 17,000 patient records in the MIMIC-III database~\cite{johnson2016mimic}. States represent clinical conditions defined by vital signs, lab values, and demographic features, while actions correspond to combinations of intravenous fluids and vasopressors at varying dosage levels. Patient survival yields a reward of +1 and death a reward of 0.
Since this benchmark remains tractable for exact model checking, we can also provide a formal analysis of the ICU-Sepsis MDP that goes beyond standard RL evaluation~\cite{choudhary2024icu}.
Using \emph{COOL-MC}, we automatically label states with clinically meaningful atomic propositions based on their state feature values, establish hard bounds on achievable outcomes, and verify policies' temporal properties and explainability via PCTL queries, providing insights into this benchmark that have not been reported.
For larger clinical MDPs where constructing the full model is infeasible, \emph{COOL-MC} can still provide such analysis by operating solely on the induced DTMC of a trained RL policy~\cite{DBLP:conf/setta/GrossJJP22}.

We first model the full MDP in the PRISM modeling language~\cite{prism_manual} and apply probabilistic model checking to establish hard bounds on achievable outcomes and characterize the inherent limitations of the disease process.
Then, we train a safe RL policy that achieves optimal survival probability and apply \emph{COOL-MC} to verify and explain its behavior on the induced DTMC through PCTL queries, feature pruning, and feature-importance permutation ranking.
This analysis reveals, for instance, that \emph{our trained policy} relies predominantly on prior dosing history rather than the patient's evolving condition, suggesting it may have learned an autoregressive heuristic that could fail to generalize beyond the training data.
Our results illustrate how \emph{COOL-MC} can provide temporal, verifiable, and interpretable analysis of treatment policies, showing its potential as a tool for trustworthy clinical decision support.

\textbf{Our main contribution} is the demonstration of \emph{COOL-MC} as a tool for healthcare applications, combining probabilistic model checking with safe and explainable RL to provide the first formal analysis of the ICU-Sepsis benchmark, including hard bounds on achievable outcomes, verified optimal policies, and explanations of learned treatment policies.

\section{Related Work}
In this section, we review related work across three areas that inform our approach: AI methods for sepsis detection and treatment, probabilistic model checking as a verification framework, and methods for ensuring the safety and explainability of RL policies.

\subsection{AI for Sepsis}
The application of artificial intelligence to sepsis spans a broad spectrum, from early detection to treatment optimization.
Wu et al.\ provide a comprehensive survey of AI in clinical decision-making for sepsis~\cite{wu2021artificial}. On the detection side, machine learning models have been developed to identify sepsis onset from electronic health records, with a growing emphasis on explainability~\cite{yang2020explainable,thakur2025explainable}. On the treatment side, RL has emerged as a framework for optimizing sequential treatment decisions. The foundational work by Komorowski et al.\cite{komorowski2018artificial} formulated sepsis treatment as an MDP, discretizing patient states via clustering and treatment actions into 25 combinations of intravenous fluids and vasopressors, and applied policy iteration to learn optimal dosing strategies from the MIMIC-III database. Other work has extended this direction of using RL for sepsis treatment~\cite{yu2023towards,sun2025exploring,shi2025between,kim2025creating,tu2025offline,wang2022learning,huang2022reinforcement,liu2021offline,wu2023value,peng2018improving,choi2024deep,roggeveen2021transatlantic,nanayakkara2022unifying,wang2024clinical,zhang2024optimizing,yu2019deep,tamboli2024reinforcedsequentialdecisionmakingsepsis}. Choudhary et al.~\cite{choudhary2024icu} distilled this formulation into the ICU-Sepsis benchmark, a standardized MDP environment that enables reproducible evaluation of RL algorithms for sepsis treatment. In this work, we focus on this benchmark as our case study.

A central challenge in applying RL to healthcare is evaluating learned policies without deploying them on real patients. Komorowski et al.~\cite{komorowski2018artificial} addressed this by using off-policy evaluation with weighted importance sampling, which provides statistical estimates of a policy's value from observational data. While effective, such estimates carry inherent variance and offer no formal guarantees.
Our work takes a fundamentally different approach: rather than statistically estimating policy value, we use probabilistic model checking to formally verify properties of the learned policy on its induced DTMC, providing exact probabilities.

Jia et al. use safety engineering processes (hazard analysis, safety cases) to identify and mitigate unsafe RL behavior before deployment~\cite{jia2020safe}, whereas our work uses probabilistic model checking with PCTL to formally verify properties of learned policies.

\subsection{Probabilistic Model Checking}
Probabilistic model checking is a well-established technique for verifying properties of stochastic systems~\cite{baier2008principles,baier2018model,DBLP:conf/birthday/KwiatkowskaN025}. It has been applied in various domains, including healthcare~\cite{chen2015data}, but its use in the context of RL for sepsis treatment has not been previously explored. To enable model checking of the ICU-Sepsis benchmark, we encode it in the PRISM modeling language~\cite{prism_manual}. For larger MDPs where constructing the full model is infeasible, \emph{COOL-MC}~\cite{DBLP:conf/setta/GrossJJP22} provides a scalable alternative by constructing only the reachable state space induced by a trained policy, enabling verification of PCTL properties on a reduced model. While the ICU-Sepsis MDP is tractable for exact model checking, we use it as a case study for \emph{COOL-MC} to demonstrate and validate its potential application in healthcare.
\emph{COOL-MC} uses internally the probabilistic model checker \emph{Storm}~\cite{DBLP:journals/sttt/HenselJKQV22}.

\subsection{Safe and Explainable RL}
Ensuring the safety, reliability, and interpretability of RL systems is essential for healthcare deployment~\cite{habli2020artificial}.
There exists a variety of related work focusing on the trained RL policy verification~\cite{DBLP:conf/sigcomm/EliyahuKKS21,DBLP:conf/sigcomm/KazakBKS19,DBLP:journals/corr/DragerFK0U15,DBLP:conf/pldi/ZhuXMJ19,DBLP:conf/seke/JinWZ22}.
For instance, RL methods with shielding~\cite{DBLP:conf/aaai/AlshiekhBEKNT18,DBLP:journals/cacm/KonighoferBJJP25} restrict the agent's actions during training to enforce formal safety specifications. While shielding has been studied extensively~\cite{DBLP:journals/cacm/KonighoferBJJP25,DBLP:conf/amcc/Bastani21,DBLP:conf/hicss/McCalmonLGCHA23,DBLP:conf/aaai/Carr0JT23,DBLP:conf/ieeecai/QiuJYWZ24,DBLP:journals/cce/GeroldL26}, its application to sepsis treatment remains largely unexplored.
There is also work on safe behavioral cloning~\cite{ashok2019sos,DBLP:conf/tacas/AshokJKWWY21,azeem2025counterexample,azeem20251,DBLP:conf/qest/AshokBCKLT19}.
Other works on safe policy learning include~\cite{DBLP:conf/ijcai/WienhoftSSDB023,DBLP:conf/iclr/HogewindSK023,DBLP:conf/aaai/SimaoS023,DBLP:conf/nips/SuilenS0022,DBLP:journals/jair/BadingsRAPPSJ23}.
We use safe behavioral cloned policies as a basis for RL with shielding.

On the explainability side, understanding what drives a trained policy's decisions is critical for clinical trust. There is work that explains the trained sepsis policies~\cite{komorowski2018artificial} by using a surrogate model. We use feature pruning~\cite{DBLP:conf/esann/GrossS24} to identify which state features the policy relies on, and we apply \emph{COOL-MC}'s state labeling method to incorporate explainability outcomes into the induced Markov chain, allowing them to be queried through PCTL properties~\cite{gross2025pctl}. This combination of safe RL, formal verification, and explainability for healthcare RL is, to our knowledge, novel.

\section{Background}
First, we introduce \emph{Markov decision processes (MDPs)} as the formal framework for sequential decision-making. We then describe probabilistic model checking for verifying policy properties, behavioral cloning and safe RL for training policies, and different explainability methods for explaining their decisions.

\subsection{Probabilistic Systems}
A \textit{probability distribution} over a set $X$ is a function $\mu \colon X \rightarrow [0,1]$ with $\sum_{x \in X} \mu(x) = 1$. The set of all distributions on $X$ is $Distr(X)$.

\begin{definition}[MDP]\label{def:mdp}
A \emph{MDP} is a tuple $M = (S,s_0,Act,Tr, rew, AP,L)$
where $S$ is a finite, nonempty set of states; $s_0 \in S$ is an initial state; $Act$ is a finite set of actions; $Tr\colon S \times Act \rightarrow Distr(S)$ is a partial probability transition function and $Tr(s,a,s')$ denotes the probability of transitioning from state $s$ to state $s'$ when action $a$ is taken;
$rew \colon S \times Act \rightarrow \mathbb{R}$ is a reward~function;
$AP$ is a set of atomic propositions;
$L \colon  S \rightarrow 2^{AP}$ is a labeling~function.
\end{definition}

We represent each state $s \in S$ as a vector of $d$ features $(f_1, \dots, f_d)$, where $f_i \in \mathbb{Q}$.
The available actions in $s \in S$ are $Act(s) = \{a \in Act \mid Tr(s,a) \neq \bot\}$ where $Tr(s, a) \neq \bot$ is defined as action $a$ at state $s$ does not have a transition (action $a$ is not available in state $s$).
In our setting, we assume that all actions are available at all states.

\begin{definition}[DTMC]\label{def:dtmc}
A \emph{discrete-time Markov chain (DTMC)} is a tuple $D = (S, s_0, Tr, AP, L)$
where $S$, $s_0$, $AP$, and $L$ are as in Definition~\ref{def:mdp}, 
and $Tr \colon S \rightarrow Distr(S)$ is a probability transition~function.
\end{definition}

In many practical settings, the agent does not have direct access to the underlying state $s \in S$ of the MDP. Instead, the agent receives an observation that may represent a transformed view of the true state. We formalize this through an observation function.

\begin{definition}[Observation]
    We define the observation function $\mathbb{O} \colon S \rightarrow O$ as a function that maps a state $s \in S$ to an observation $o \in O$.
    An observation $o \in O$ is a vector of features $(f_1, \dots, f_d)$ where $f_j \in \mathbb{Q}$.
    The observed features may differ from the exact state features.
\end{definition}

A policy then operates on observations rather than on the underlying states directly. This distinction is important in our setting, where the MDP uses integer state identifiers internally but the policy receives 47-dimensional clinical feature vectors derived from cluster centroids.

\begin{definition}
    A \emph{memoryless deterministic policy $\pi$} for an MDP $M$ is a function $\pi \colon O \rightarrow Act$ that maps an observation $o \in O$ to action $a \in Act$.
\end{definition}

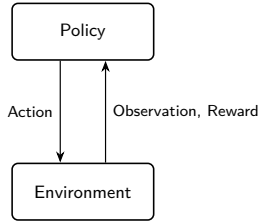
\begin{figure}[]
\centering
\scalebox{0.25}{
    \tikzset{
    process box/.style={
        rectangle,
        rounded corners=3pt,
        draw=black,
        line width=0.8pt,
        fill=white,
        minimum width=2.5cm,
        minimum height=1cm,
        align=center,
        font=\sffamily\small
    },
    arrow/.style={
        -{Stealth[length=2mm, width=1.5mm]},
        line width=0.6pt,
        color=black
    },
    label style/.style={
        font=\sffamily\footnotesize
    }
}

\begin{tikzpicture}

\node[process box] (policy) {Policy};
\node[process box, below=1.8cm of policy] (env) {Environment};

\draw[arrow] ([xshift=-0.4cm]policy.south) -- node[label style, left] {Action} ([xshift=-0.4cm]env.north);

\draw[arrow] ([xshift=0.4cm]env.north) -- node[label style, right] {Observation, Reward} ([xshift=0.4cm]policy.south);

\end{tikzpicture}
    }
\caption{Sequential decision-making loop. The agent receives an observation and reward from the environment, selects an action according to its policy, and the environment transitions to a new state.}
\label{fig:rl}
\end{figure}

Applying a policy $\pi$ to an MDP $M$ with observation function $\mathbb{O}$ yields an \emph{induced DTMC} $D^\pi$ where all non-determinism is resolved: for each state $s$, the transition function becomes $Tr(s, s') = Tr(s, \pi(\mathbb{O}(s)), s')$.
The induced DTMC fully characterizes the observable behavior of the policy: the states visited, the transitions taken, and the probabilities of all outcomes.
The interaction between the policy and environment is depicted in Figure~\ref{fig:rl}.

\subsection{Probabilistic Model Checking}

\emph{Probabilistic model checking} enables the verification of quantitative properties of stochastic systems.
\emph{COOL-MC} uses \emph{Storm}~\cite{DBLP:journals/sttt/HenselJKQV22} internally as its model checker, which can verify properties of both MDPs and DTMCs.
Among the most fundamental properties are \emph{reachability} queries, which assess the probability of a system reaching a particular state.
For example, one might say: ``The reachability probability of reaching an unsafe state is~0.1.''

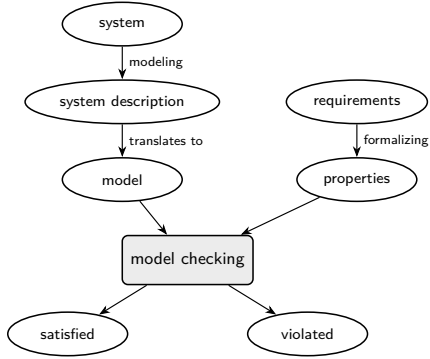
\begin{figure}[htbp]
    \centering
    \scalebox{0.4}{
    \tikzset{
    ellipse node/.style={
        ellipse,
        draw=black,
        line width=0.8pt,
        fill=white,
        minimum width=2.2cm,
        minimum height=0.8cm,
        align=center,
        font=\sffamily\footnotesize
    },
    process block/.style={
        rectangle,
        rounded corners=3pt,
        draw=black,
        line width=0.8pt,
        fill=gray!15,
        minimum width=2.4cm,
        minimum height=0.9cm,
        align=center,
        font=\sffamily\small
    },
    arrow/.style={
        -{Stealth[length=2mm, width=1.5mm]},
        line width=0.6pt,
        color=black
    },
    label style/.style={
        font=\sffamily\scriptsize
    }
}

\begin{tikzpicture}[node distance=0.7cm]

\node[ellipse node] (system) {system};
\node[ellipse node, below=0.6cm of system] (systemdesc) {system description};
\node[ellipse node, below=0.6cm of systemdesc] (model) {model};

\node[ellipse node, right=1.2cm of systemdesc] (requirements) {requirements};
\node[ellipse node, below=0.6cm of requirements] (properties) {properties};

\node[process block, below=0.6cm of model, xshift=1.2cm] (modelchecking) {model checking};

\node[ellipse node, below left=0.6cm and 0.2cm of modelchecking] (satisfied) {satisfied};
\node[ellipse node, below right=0.6cm and 0.2cm of modelchecking] (violated) {violated};

\draw[arrow] (system) -- node[label style, right] {modeling} (systemdesc);
\draw[arrow] (systemdesc) -- node[label style, right] {translates to} (model);
\draw[arrow] (requirements) -- node[label style, right] {formalizing} (properties);
\draw[arrow] (model) -- (modelchecking);
\draw[arrow] (properties) -- (modelchecking);
\draw[arrow] (modelchecking) -- (satisfied);
\draw[arrow] (modelchecking) -- (violated);

\end{tikzpicture}
    }
    \caption{General model checking workflow~\cite{DBLP:journals/sttt/HenselJKQV22}. The system is formally modeled, the requirements are formalized, and both are input to a model checker such as \emph{Storm}, which verifies the property.}
    \label{fig:model_checking}
\end{figure}

The general workflow for model checking is as follows (see also Figure~\ref{fig:model_checking}).
First, the system is formally modeled using a language such as PRISM~\cite{prism_manual}.
Next, the property of interest is formalized in a temporal logic.
Using these inputs, the model checker verifies whether the property holds or computes the relevant probability.

The PRISM language~\cite{prism_manual} describes probabilistic systems as a collection of \emph{modules}, each containing typed \emph{variables} and guarded \emph{commands}.
A state of the system is a valuation of all variables, and the state space $S$ is the set of all possible valuations.
Each command takes the~form
\[
\texttt{[action]}\ g \rightarrow \lambda_1 : u_1 + \cdots + \lambda_n : u_n
\]
where $g$ is a Boolean guard over the variables, each $\lambda_j$ is a probability, and each $u_j$ is an update that assigns new values to the variables.
When the system is in a state satisfying $g$, the command can be executed: update $u_j$ is applied with probability $\lambda_j$, transitioning the system to a new state.
For an MDP, each enabled command in a state corresponds to a distinct probabilistic choice, and the nondeterministic selection among enabled commands is resolved by a policy.
Listing~\ref{lst:prism_example} shows a simplified excerpt of the sepsis treatment MDP in PRISM.
The model declares a single module with a state variable \texttt{s} ranging over integer identifiers.
Each command specifies a transition: a guard on the current state, an action label, and a probabilistic update.
For instance, from state $s=0$, taking action \texttt{a0} leads to state $s=1$ with probability $0.7$ and to state $s=2$ with probability $0.3$.
States $s=2$ and $s=3$ represent absorbing terminal states corresponding to survival and death, respectively.
A reward structure assigns a reward of $+1$ to the survival state.
Additionally, PRISM supports \emph{atomic labels} that annotate states with propositions.
These labels are defined using \texttt{label} declarations, which assign a name to a Boolean condition over the state variables.
For example, \texttt{label "survival" = s=2;} marks all states where $s=2$ with the proposition \textit{survival}.
Such labels serve as the atomic propositions referenced in PCTL properties. \emph{COOL-MC} leverages this mechanism to label states with clinically meaningful propositions derived from the 47-dimensional feature vectors (e.g., \textit{high\_sofa}).
\emph{COOL-MC} allows automated user-specified state labeling of PRISM models.
For a complete formal treatment of the PRISM semantics, we refer to~\cite{prism_manual}.

\begin{figure}[htbp]
\begin{lstlisting}[
    language={},
    basicstyle=\ttfamily\small,
    keywordstyle=\bfseries,
    commentstyle=\itshape\color{gray},
    frame=single,
    numbers=left,
    numberstyle=\tiny,
    caption={Simplified PRISM model of the sepsis treatment MDP. States are encoded as integers; transitions are guarded by the current state and labeled with actions. Atomic labels annotate states with propositions used in PCTL~properties.},
    label={lst:prism_example},
    escapeinside={(*}{*)}
]
mdp

module sepsis
  s : [0..3] init 0; // state variable

  // From state 0, action a0:
  //   -> state 1 (w.p. 0.7), state 2 (w.p. 0.3)
  [a0] s=0 -> 0.7:(s'=1) + 0.3:(s'=2);

  // From state 0, action a1:
  //   -> state 1 (w.p. 0.4), state 3 (w.p. 0.6)
  [a1] s=0 -> 0.4:(s'=1) + 0.6:(s'=3);

  // From state 1, action a0:
  //   -> state 2 (w.p. 0.8), state 3 (w.p. 0.2)
  [a0] s=1 -> 0.8:(s'=2) + 0.2:(s'=3);

  // Absorbing states (survival and death)
  [end] s=2 -> 1.0:(s'=2); // survival
  [end] s=3 -> 1.0:(s'=3); // death

endmodule

// Atomic labels for state propositions
label "survival" = s=2;
label "death" = s=3;

// Reward: +1 for reaching survival state
rewards
  s=2 : 1;
endrewards
\end{lstlisting}
\end{figure}

Properties are specified using PCTL~\cite{hansson1994logic}, a branching-time temporal logic for reasoning about probabilities over paths.
In this paper, we use two path operators.
The \emph{eventually} operator $\lozenge\,\varphi$ states that $\varphi$ holds at some future state along a path.
The \emph{until} operator $\varphi_1 \;\mathcal{U}\; \varphi_2$ states that $\varphi_1$ holds at every state along a path until a state is reached where $\varphi_2$ holds.
A PCTL property has the form
\[
P_{\sim p} (\psi)
\]
where $\psi$ is a path formula (such as $\lozenge\,\varphi$ or $\varphi_1 \;\mathcal{U}\; \varphi_2$), $\sim$ is a comparison operator ($<$, $\leq$, $\geq$, $>$), and $p \in [0,1]$ is a probability threshold.
Beyond checking whether a property is satisfied, \emph{Storm} can compute the exact probability, denoted $P_{=?}(\psi)$.
We refer to~\cite{hansson1994logic,baier2008principles} for a detailed description and additional PCTL operators.

For MDPs, \emph{Storm} can additionally synthesize optimal policies.
Given an MDP and a property specification, \emph{Storm} computes a policy that maximizes or minimizes the probability of satisfying the property, denoted $P_{\max}(\psi)$ and $P_{\min}(\psi)$ respectively.
For instance, $P_{\max}(\lozenge\,\textit{survival})$ yields the maximum survival probability achievable under any policy, along with the policy that achieves it.
This capability allows us to extract an optimal policy for a specified property, which can then serve as the expert dataset for behavioral cloning~\cite{DBLP:conf/setta/GrossJJP22}.

\subsection{Data-Driven Policy Learning}
In this section, we introduce behavioral cloning and safe RL.
\subsubsection{Behavioral cloning}
Behavioral cloning is an imitation learning approach that frames policy learning as supervised classification~\cite{DBLP:journals/corr/abs-2204-05618}.
Given a dataset of expert demonstrations consisting of observation-action pairs $\{(\mathbb{O}(s), \pi^*(\mathbb{O}(s)))\}$, behavioral cloning trains a policy $\pi$ to predict the expert's action for each observation.
A key advantage of behavioral cloning is its simplicity: it reduces sequential decision-making to supervised learning.
However, it assumes access to high-quality expert demonstrations and can suffer from \emph{distribution shift}, where errors compound as the learned policy visits states not represented in the training data~\cite{kumar2022should}.
To explore the state and action space in greater detail, safe RL can be a viable alternative.

\subsubsection{Safe RL}
The standard goal of RL is to learn a policy $\pi$ for an MDP that maximizes the expected accumulated discounted reward~\cite{DBLP:journals/ml/Bekkemoen24}:
\[
\mathbb{E}\left[\sum_{t=0}^{N} \gamma^t \cdot rew(s_t, a_t)\right],
\]
where $\gamma \in [0,1]$ is the discount factor, $rew(s_t, a_t)$ is the reward at time step $t$, and $N$ is the episode length.

A key challenge in applying RL to safety-critical domains is that it offers no formal guarantees on constraint satisfaction during or after training.
\emph{Shielding}~\cite{DBLP:conf/aaai/AlshiekhBEKNT18} addresses this by composing the learning agent with a reactive safety mechanism that monitors and, when necessary, overrides the agent's actions to enforce formal safety specifications.

In \emph{post-shielding}~\cite{DBLP:journals/cacm/KonighoferBJJP25}, the RL agent selects actions freely according to its policy $\pi$, and the shield intervenes only if the chosen action would violate the safety specification.
At each time step, the shield evaluates whether the agent's proposed action $a_t = \pi(\mathbb{O}(s_t))$ belongs to the set of safe actions $Act_{\text{safe}}(s_t) \subseteq Act$.
If so, the action is executed; otherwise, the shield replaces it with a safe alternative.
The shielded policy is:
\begin{equation}
    \pi_{\text{safe}}(s_t) = \begin{cases} \pi(\mathbb{O}(s_t)) & \text{if } \pi(\mathbb{O}(s_t)) \in Act_{\text{safe}}(s_t) \\ \operatorname{correct}(\pi(\mathbb{O}(s_t)), Act_{\text{safe}}(s_t)) & \text{otherwise} \end{cases}
\end{equation}
where $\operatorname{correct}$ selects a valid action.
This allows the agent to explore and optimize freely while the shield guarantees that no unsafe action is ever executed.

\subsection{Explainability Methods}\label{sec:xrl}
Explainability methods aim to make trained RL policies understandable~\cite{DBLP:journals/csur/MilaniTVF24}.
Global methods interpret the overall policy's behavior~\cite{DBLP:conf/aiide/SieusahaiG21}, while local methods explain a policy's decision-making in individual states.

In this work, we use two explainability methods: \emph{feature pruning} and \emph{feature permutation importance} in combination with PCTL model checking.

Consider a neural network policy operating on observations with $d$ input features and $|Act|$ output actions, encoding a function $g \colon \mathbb{Q}^d \to \mathbb{R}^{|Act|}$.
The function $g$ is parameterized by a sequence of weight matrices $\vec{W}^{(1)}, \dots, \vec{W}^{(k)}$ with $\vec{W}^{(\ell)} \in \mathbb{R}^{d_\ell \times d_{\ell-1}}$ for $\ell = 1, \dots, k$.
Feature pruning eliminates all outgoing connections from a particular input neuron by setting $\vec{W}^{(1)}_{j,i} = 0$ for all $j$, effectively removing the $i$-th input feature $f_i$ from the policy's decision-making~\cite{DBLP:conf/esann/GrossS24}.
By measuring the resulting change in the policy's behavior (e.g., degradation in survival probability on the induced DTMC), we quantify the importance of each feature.

Feature permutation importance~\cite{breiman2001random} offers a complementary, model-agnostic measure of feature relevance.
For a given observation $o = \mathbb{O}(s)$ with features $(f_1, \dots, f_d)$, the values of a single feature $f_i$ are randomly permuted across observations while all other features remain fixed.
The policy is then re-evaluated on the permuted inputs, and the change in action selection is measured.
A large change indicates that the policy's decisions depend strongly on $f_i$.
Unlike feature pruning, which permanently modifies the network weights, permutation importance leaves the policy intact. It can be computed per state, providing a local importance ranking that can be used to label states in the induced DTMC for further PCTL analysis.

\section{Methodology}
Our methodology proceeds in four stages.
First, we encode the ICU-Sepsis MDP~\cite{choudhary2024icu} in the PRISM modeling language and incrementally build the full model, labeling states with clinically meaningful atomic propositions via COOL-MC (\S\ref{subsec:mdp-encoding}).
Second, we perform probabilistic model checking on the full MDP to verify properties, synthesize optimal policies, and characterize admissible treatment trajectories (\S\ref{subsec:model-checking}).
Third, we train deep RL policies using behavioral cloning from the verified optimal policy as a warm-start, followed by RL with post-shielding to safely explore alternative strategies (\S\ref{subsec:safe-rl}).
Fourth, we verify each trained policy on its induced DTMC, applying PCTL queries and feature pruning to assess safety guarantees and create explanations of the policy's decision-making (\S\ref{subsec:policy-verification}).
Finally, we describe the limitations of our methodology (\S\ref{subsec:limitations}).

\subsection{MDP Encoding}\label{subsec:mdp-encoding}
We encode the ICU-Sepsis MDP proposed by Choudhary et al.~\cite{choudhary2024icu} in the PRISM modeling language. This MDP is derived from approximately 17,000 sepsis patients in the MIMIC-III v1.4 database~\cite{PhysioNet-mimiciii-1.4}, following the data processing pipeline of Komorowski et al.~\cite{komorowski2018artificial}.
The model comprises 717 states and 25 actions. Of these, 716 states represent clinical patient conditions, while the additional state $s_0$ serves as an auxiliary initial state that transitions probabilistically to the clinical states according to the patient admission distribution observed in the dataset. This ensures that model checking results reflect the full population of incoming patients rather than a single fixed starting condition.
States are labeled with atomic propositions (e.g., \textit{survived}, \textit{died}) to enable PCTL property specification.

Each of the 716 clinical states $s \in S$ corresponds to a cluster of patient conditions, characterized by a 47-dimensional feature vector $(f_1, \dots, f_{47})$ representing the cluster centroid.
The 47 features span demographics, vitals, blood chemistry and metabolic markers, liver and coagulation indicators, hematology, blood gas values, organ dysfunction and severity scores, fluid balance, vasopressor dosing, and ventilation status~\cite{choudhary2024icu}. 

\begin{figure}[t]
\centering
\begin{tikzpicture}[
    box/.style={draw, rounded corners, minimum width=2cm, minimum height=0.9cm, align=center, font=\small},
    featurebox/.style={draw, rounded corners, minimum width=3.3cm, align=left, font=\scriptsize, inner sep=6pt},
    processbox/.style={draw, rounded corners, fill=orange!15, minimum width=2.2cm, minimum height=0.9cm, align=center, font=\small},
    arrow/.style={->, thick, >=stealth},
    label/.style={font=\footnotesize\bfseries, align=center}
]

\draw[draw=black!60, dashed, rounded corners, fill=gray!5] (-5.5, -3.2) rectangle (7.5, 4.0);
\node[font=\small\bfseries, anchor=north west] at (-5.3, 3.9) {COOL-MC};

\node[label] at (-3.2, 3.0) {Internal (PRISM)};
\node[box, fill=blue!10, minimum width=1.8cm] (prism_state) at (-3.2, 2.0) {$id = 42$};
\node[font=\scriptsize, text=gray, align=center] at (-3.2, 1.1) {State s};

\draw[arrow, thick, color=black!70] (-1.5, 2.0) -- node[above, font=\scriptsize] {$\mathbb{O}(s)$} (0.3, 2.0);

\node[featurebox, fill=green!10] (inflated) at (3.0, 2.0) {
    \textbf{$o = \mathbb{O}(s) = (f_1, \dots, f_{47})$}\\[3pt]
    $f_1$: Age = 67\\
    $f_2$: Heart Rate = 98\\
    $f_3$: SpO2 = 94\\
    $f_4$: Lactate = 3.1\\
    \quad$\vdots$\\
    $f_{47}$: Ventilation = 1
};

\node[processbox] (policy) at (-0.5, -1.7) {Policy $\pi(o)$};
\node[processbox] (explain) at (3.0, -1.7) {Explainability};
\node[processbox] (labeling) at (6.2, -1.7) {State Labeling};

\draw[arrow] (inflated.south) -- ++(0, -0.5) -| (policy.north);
\draw[arrow] (inflated.south) -- ++(0, -0.5) -| (explain.north);
\draw[arrow] (inflated.south) -- ++(0, -0.5) -| (labeling.north);

\node[font=\scriptsize, text=gray, align=center] at (-0.5, -2.6) {Action\\selection};
\node[font=\scriptsize, text=gray, align=center] at (3.0, -2.6) {Feature permutation\\ranking, ...};
\node[font=\scriptsize, text=gray, align=center] at (6.2, -2.6) {Feature-based\\state annotation, ...};

\end{tikzpicture}
\caption{State handling example within COOL-MC. Internally, COOL-MC uses PRISM's integer state encoding for model checking. Whenever a state is passed to the policy, an explainability method, or state labeling, the observation function $\mathbb{O}$ maps it to its corresponding 47-dimensional observation, enabling decisions, analyses, and annotations based on clinically meaningful patient characteristics.}
\label{fig:state-representation}
\end{figure}
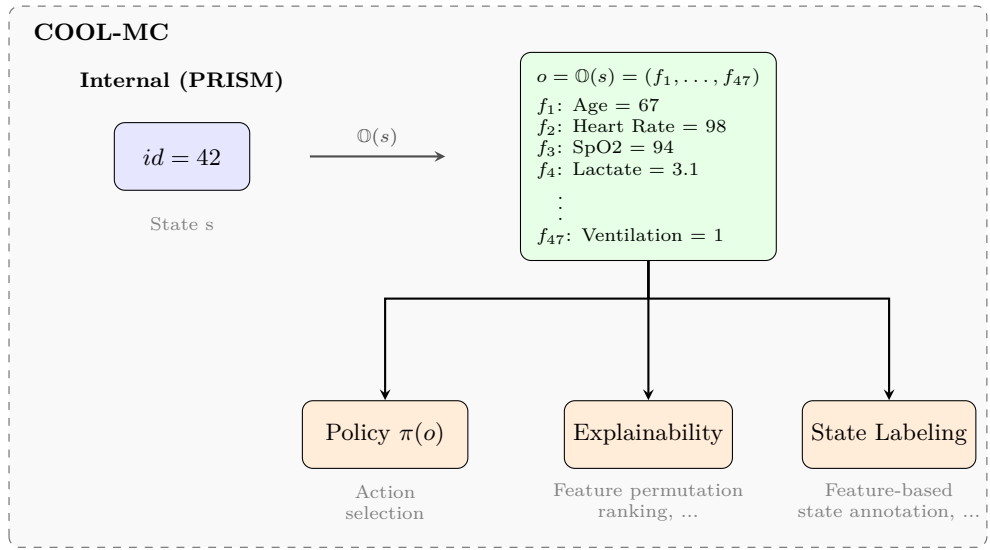

While PRISM encodes states solely as integer identifiers, \emph{COOL-MC} applies the observation function $\mathbb{O}$ to map each integer state to its corresponding 47-dimensional observation whenever the state is passed to the policy, an explainability method, or state labeling.
This enables the policy to reason over clinically meaningful patient characteristics rather than abstract state indices, and allows explainability and state-labeling techniques to analyze and annotate states based on clinical observations (see Figure~\ref{fig:state-representation}).

The 25 actions $a \in Act$ represent all combinations of two treatment interventions: \emph{intravenous (IV) fluids} and \emph{vasopressors}. Following Choudhary et al.~\cite{choudhary2024icu}, the dosage of each intervention is discretized into 5 levels (including a zero-dose level), yielding $5 \times 5 = 25$ possible treatment combinations per time step.
The discretization bins range from no administration to high-dose regimens, capturing the clinically relevant dosage spectrum.

The reward structure is sparse and binary: all intermediate rewards are $0$, and at the end of each episode, the agent receives $rew = +1$ if the patient survives (based on 90-day mortality) and $rew = 0$ if the patient dies.

In summary, the encoded PRISM MDP with the observation function is formally defined as:
\begin{gather*}
S = \{0, 1, \ldots, 714\} \cup \{s_{\text{init}}, s_{\text{survived}}, s_{\text{died}}\} \label{eq:sepsis} \\
\mathbb{O}(s) = (age, \dots, sofa) \\
Act = \{0, 1, \ldots, 24\} \quad \text{(5 IV fluid levels} \times \text{5 vasopressor levels)} \notag \\
rew(s, a, s') = \begin{cases}
        +1 & \text{if } s' = s_{\text{survived}} \\
        0 & \text{otherwise}
        \end{cases} \notag
\end{gather*}

\subsection{Probabilistic Model Checking}\label{subsec:model-checking}
Using \emph{COOL-MC}, we construct and label the full MDP and perform model checking via \emph{Storm} to (a) verify PCTL properties such as maximum and minimum survival probabilities, (b) synthesize optimal policies, and (c) characterize admissible treatment trajectories.
The observation-action pairs of the optimal policy then serve as expert demonstrations for behavioral cloning.

During the MDP construction, we label states with user-specified atomic propositions conditioned on their feature values.
This labeling step is essential because the model checker operates over abstract state identifiers (see Figure~\ref{fig:state-representation}) and does not have direct access to the 47-dimensional feature vectors. States whose feature assignments fall within clinically meaningful ranges are labeled accordingly.
For example, a state whose sofa feature indicates value above the 75\%-percentile is labeled as \textit{high\_sofa}, allowing verification of properties such as $P_{\max}(\neg\textit{high\_sofa} \;\mathcal{U}\; \textit{survived})$: the maximum probability of reaching survival without ever passing through a high sofa state.

\subsection{Safe RL Policy Training}\label{subsec:safe-rl}
We train deep RL policies on the ICU-Sepsis MDP in two phases. In the first phase, we perform \emph{behavioral cloning} from the optimal policy $\pi^*$ synthesized via probabilistic model checking, using observation-action pairs $\{(\mathbb{O}(s), \pi^*(s))\}$ as expert demonstrations to pretrain the agent~\cite{azeem20251}.
This gives the agent an initial policy that is already close to optimal.

In the second phase, we fine-tune the policy using \emph{RL with post-shielding}~\cite{DBLP:journals/cacm/KonighoferBJJP25}.
This allows the agent to explore actions beyond those prescribed by the cloned policy, potentially discovering alternative optimal actions.
At each state $s_t$, the agent selects an action according to its current policy; a shield then verifies whether this action belongs to the set of safe actions $Act_{\text{safe}}(s_t)$ derived from the safety specification.
If the selected action is safe, it is executed; otherwise, the shield replaces it with a safe alternative.
This mechanism ensures that no action violating the safety specification is ever executed, while still allowing the agent to explore the space of safe behaviors.

While \emph{Storm} can directly synthesize an optimal policy via model checking~\cite{DBLP:journals/sttt/HenselJKQV22}, this approach has several limitations. First, the synthesized policy operates on integer state identifiers and cannot generalize beyond the specific MDP for which it was computed.
By training a policy using behavioral cloning and RL, we obtain a policy that operates in the 47-dimensional feature space and can potentially generalize to unseen patient conditions.
Second, for larger MDPs, exact policy synthesis via model checking becomes infeasible due to the state explosion problem. In such cases, training a policy on observations from a smaller MDP and verifying it on the induced DTMC of the original MDP~\cite{azeem20251}, which resolves all nondeterminism, is significantly more tractable than exhaustive analysis of the full MDP, while still providing meaningful guarantees about the policy's behavior~\cite{DBLP:conf/setta/GrossJJP22}.
Finally, the trained policy is amenable to the explainability methods provided by COOL-MC, such as feature pruning~\cite{DBLP:conf/esann/GrossS24} and feature-importance permutation ranking~\cite{gross2025pctl,breiman2001random}.
Training an RL policy thus bridges the gap between the formal guarantees of model checking and scalability.

\begin{algorithm}[t]
\caption{COOL-MC: Formal Verification of Policies}
\label{alg:qverifier}
\begin{algorithmic}[1]
\Require MDP $M$, trained policy $\pi$,  PCTL property $\varphi$, observation function $\mathbb{O}$, \textsc{Label}(o)
\Ensure Satisfaction result and probability $p$

\Statex
\Statex \textbf{Stage 1: Induced DTMC Construction}
\State $D^\pi \gets (S^\pi, s_0, Tr^\pi, AP, L)$ where $S^\pi \gets \emptyset$, $Tr^\pi \gets \emptyset$
\State \textsc{BuildDTMC}($s_0$)

\Statex
\Statex \textbf{Stage 2: Probabilistic Model Checking}
\State $(result, p) \gets \emph{Storm}.\text{verify}(D^{\pi}, \varphi)$
\State \Return $(result, p)$

\Statex
\Procedure{BuildDTMC}{$s$}
    \If{$s \in S^\pi$ \textbf{or} $s$ is not relevant for $\varphi$}
        \State \Return
    \EndIf
    \State $S^\pi \gets S^\pi \cup \{s\}$
    \State $L^\pi(s) \gets \textsc{Label}(\mathbb{O}(s))$
    \State $a \gets \pi(\mathbb{O}(s))$
    \ForAll{$s' \in S$ where $Tr(s, a, s') > 0$}
        \State $Tr^\pi(s, s') \gets Tr(s, a, s')$
        \State \textsc{BuildDTMC}($s'$)
    \EndFor
\EndProcedure

\end{algorithmic}
\end{algorithm}

\subsection{Policy Verification and Explanation}\label{subsec:policy-verification}
For each trained policy $\pi$, \emph{COOL-MC} constructs the induced DTMC $D^\pi$ by exploring only the reachable state space under $\pi$ (see Algorithm~\ref{alg:qverifier}).
Starting from the initial state $s_0$, the construction proceeds as a depth-first traversal of the state space.
At each visited state $s$, the policy selects an action $a = \pi(\mathbb{O}(s))$, and all successor states $s'$ with $Tr(s, a, s') > 0$ are added to the induced DTMC along with their transition probabilities.
The procedure recurses into each successor that has not yet been visited and is relevant for the property $\varphi$ under verification.
During this traversal, each state is also labeled with the user-specified labeler \textsc{Label}(o), enabling the formulation and verification of domain-specific properties.
This incremental approach offers two key advantages.
First, it constructs only the fragment of the full MDP that is actually reachable under $\pi$, which can be substantially smaller than the complete state space, mitigating the state explosion problem.
Second, it produces a DTMC rather than an MDP, since all nondeterminism is resolved by $\pi$, making model checking significantly more efficient.
Once the induced DTMC is fully constructed, it is passed to \emph{Storm} for probabilistic model checking.
On this reduced model, we perform several analyses.

First, we verify PCTL properties to assess policy-specific outcomes, answering questions such as the survival probability under $\pi$ or whether certain patient subgroups face disproportionately low survival rates.

Second, we combine verification with the state labeling approach described in \S\ref{subsec:model-checking}. By labeling states in the induced DTMC with clinically meaningful propositions (e.g., \textit{high\_sofa}), we can characterize treatment trajectories and identify interesting states~\cite{gross2025pctl}.
Additionally, we apply feature-importance permutation ranking to identify which state features the policy relies on most across different treatment trajectories.

Third, we apply \emph{feature pruning} to the trained neural network policy~\cite{DBLP:conf/esann/GrossS24}.
By removing individual features $f_i$ and re-verifying the induced DTMC, we measure the resulting change in survival probability, quantifying the importance of each feature to the trained policy's decision-making.
For example, pruning the age-related features reveals whether the policy's decisions depend critically on the patient's age.

\subsection{Limitations}\label{subsec:limitations}
Several limitations should be noted.
First, the ICU-Sepsis MDP is an environment with 716 (+1) discrete states derived from clustering, which inevitably loses clinical granularity compared to the continuous patient feature space.
Policies verified on this MDP may not directly transfer to real clinical settings.
The transition probabilities are estimated from the MIMIC-III dataset, which reflects historical treatment practices from 2001--2012; shifts in clinical standards may limit the relevance of the learned policies.

When considering COOL-MC transfer to other MDPs for sepsis treatment, additional limitations of \emph{COOL-MC} must be taken into account.
\emph{COOL-MC} requires a discrete action space, which may necessitate further discretization when applied to domains with continuous treatment dosages~\cite{huang2022reinforcement}.
Moreover, while \emph{COOL-MC} mitigates state-space explosion through incremental model construction, sufficiently large state spaces will still face scalability issues.
Furthermore, \emph{COOL-MC} supports memoryless (Markovian) policies, meaning that treatment strategies depending on patient history rather than the current state alone cannot be directly represented or verified.
This limitation could be addressed by techniques such as state feature extension, which encodes relevant historical information into the current state representation (increasing the number of states and transitions).

As emphasized by Choudhary et al.~\cite{choudhary2024icu}, the ICU-Sepsis environment is intended as an RL benchmark and should not be used to draw conclusions that guide medical practice.

\section{Experiments}
We apply \emph{COOL-MC} to the ICU-Sepsis MDP in two parts.
Part A analyzes the full MDP to determine what no treatment policy can change: by computing best-case and worst-case probabilities across all possible policies, we characterize the theoretical bounds on achievable outcomes and identify clinical deterioration inherent to the disease process itself.
Part B then examines what the learned policy actually does: we construct the induced DTMC of a trained safe RL policy and analyze its behavior through PCTL verification, state labeling, feature pruning, and feature-importance permutation ranking.
Together, these two parts illustrate how COOL-MC bridges the gap between global bounds on any policy and a detailed, interpretable analysis of a specific learned policy.

\subsection{Setup}
We executed our experiments in a Docker container with 16\,GB RAM and an AMD Ryzen 7 7735HS processor, running Ubuntu 20.04.5 LTS.
For model checking, we use \emph{Storm} 1.7.1 (dev).
Implementation details are provided in the accompanying source code at \url{https://github.com/LAVA-LAB/COOL-MC/tree/sepsis}.

\subsection{Analysis}
Having described the experimental setup, we now present our findings. Part A reports the results of model checking the full MDP, and Part B analyzes the trained policy on its induced DTMC.

\subsubsection{Part A: What no treatment policy can change}
This section explores the theoretical limits of what any treatment policy can achieve in the modeled MDP.
The goal is to demonstrate the analytical capabilities of probabilistic model checking in a healthcare setting, not to derive clinical conclusions, as medical interpretation is outside our domain of expertise.

By model-checking the full MDP, we obtain hard bounds on what any policy can achieve.
The maximum survival probability achievable under any policy is $P_{\max}(\lozenge\;\textit{survived}) = 0.8751$ (2max), while the minimum is $P_{\min}(\lozenge\;\textit{survived}) = 0.633$ (2min).

\begin{table}[ht]
\centering
\caption{Full MDP analysis: best-case and worst-case probabilities over all possible treatment policies. Labels, such as \textit{high\_sofa}, are atomic propositions assigned to states.}
\label{tab:severity_subgroup}
\small
\begin{tabular}{lp{5.2cm}lr}
\toprule
\textbf{ID} & \textbf{Description} & \textbf{PCTL Property} & \textbf{Prob.} \\
\midrule
2max & Best achievable survival & $P_{\max}(\lozenge\ \texttt{survived})$ & 0.8751 \\
2min & Worst-case survival & $P_{\min}(\lozenge\ \texttt{survived})$ & 0.633 \\
2a & Best clean survival (no severe organ failure) & $P_{\max}(\ \neg\texttt{high\_sofa}\ \mathcal{U}\ \texttt{survived})$ & 0.4667 \\
2b & Worst clean survival (no severe organ failure) & $P_{\min}(\ \neg\texttt{high\_sofa}\ \mathcal{U}\ \texttt{survived})$ & 0.2986 \\
2c & Min.\ deaths without organ failure & $P_{\min}(\ \neg\texttt{high\_sofa}\ \mathcal{U}\ \texttt{died})$ & 0.0504 \\
2d & Unavoidable low blood pressure & $P_{\min}(\lozenge\ \texttt{low blood pressure})$ & 0.5274 \\
2e & Worst combined organ \& metabolic distress & $P_{\max}(\lozenge\ (\texttt{high\_sofa} \wedge \texttt{high\_lactate}))$ & 0.4145 \\
2f & Min.\ combined organ \& metabolic distress & $P_{\min}(\lozenge\ (\texttt{high\_sofa} \wedge \texttt{high\_lactate}))$ & 0.2047 \\
2g & Unavoidable organ dysfunction & $P_{\min}(\lozenge\ \texttt{high\_sofa})$ & 0.4428 \\
2h & Unavoidable septic shock & $P_{\min}(\lozenge\ (\texttt{high\_sofa} \wedge \texttt{low blood pressure}))$ & 0.1865 \\
2i & Min.\ deaths without hypotension & $P_{\min}(\ \neg\texttt{low blood pressure}\ \mathcal{U}\ \texttt{died})$ & 0.0532 \\
\bottomrule
\end{tabular}
\end{table}

\emph{Surviving without severe organ dysfunction is possible.}
The best any policy can achieve for clean survival, defined as reaching survival without ever entering the top 25\% SOFA states, is 0.4667 (2a), while the worst case drops to 0.2986 (2b).

\emph{Some deaths occur without severe organ failure.}
The minimum probability of dying without ever reaching a high SOFA score is 0.0504 (2c), meaning that even under the most favorable policy, at least 5.04\% of patients die without ever entering a high-SOFA state.

\emph{Low blood pressure states are unavoidable.} No policy can reduce the probability of patients reaching states with MeanBP below the 25th percentile below 0.5274 (2d).

\emph{Combined critical states are partly inevitable.}
No policy can reduce the probability of reaching both high SOFA and high lactate states (top 25\% values) below 0.2047 (2f), while under the worst policy this probability rises to 0.4145 (2e). This means treatment decisions can narrow this risk by approximately 0.21, but no policy can eliminate it.

\emph{Severe organ dysfunction itself is largely unavoidable.}
Even under the best possible policy, the probability of reaching high SOFA states cannot be reduced below 0.4428 (2g).

\emph{Combined SOFA and hypotension has a lower floor.}
No policy can reduce the probability of reaching both high SOFA and low MeanBP states below 0.1865 (2h). Furthermore, even in the absence of low MeanBP, death occurs with a probability of at least 0.0532 (2i).

\emph{Summary.}
Treatment decisions matter substantially, as the gaps between best- and worst-case survival probabilities are meaningful.
This sets realistic expectations for what any RL policy can achieve when evaluated against these theoretical bounds.

\subsubsection{Part B: What the learned policy actually does}
We train a PPO agent on the ICU sepsis MDP using a three-stage pipeline.
First, we perform \emph{behavioral cloning} pre-training for 65 epochs,
where the policy network is trained via supervised learning on observation-action pairs
extracted from the optimal scheduler of the PRISM model with respect to the
property $P_{\max}(\lozenge\;\texttt{survived})$.
The policy network consists of four fully connected layers, each with 512 neurons.
Second, the pre-trained policy is fine-tuned using PPO with a learning rate
over 25{,}000 episodes (learning rate $\alpha = 3 \times 10^{-4}$,
discount factor $\gamma = 0.99$, batch size 32, clipping parameter
$\epsilon = 0.2$).
During RL training, a \emph{post-shielding} mechanism acts as a safety layer:
Before each environment step, the agent's chosen action is checked against the
set of all equally optimal actions derived from the PRISM model; if the action
is suboptimal, a randomly selected optimal action replaces it.
This shielded training ensures that the agent only experiences optimal
trajectories.

The trained PPO policy $\pi$ was evaluated by constructing its induced DTMC $D^\pi$ and applying probabilistic model checking.
The induced DTMC comprises 717 reachable states and 35,045 transitions, compared to the full MDP's 717 states and 112,821 transitions.
The survival probability under $\pi$ is $P_{=?}(\lozenge\;\textit{survived}) = 0.8751$, matching the theoretical optimum.

We label each state with the action selected by $\pi$, using the notation $\texttt{fl}_i\texttt{vp}_j$ where $i \in \{0,\dots,4\}$ is the IV-fluid level and $j \in \{0,\dots,4\}$ the vasopressor level ($0$ = none, $4$ = maximum dose).
Across the 717 reachable states, the most frequently assigned action is $\texttt{fl}_0\texttt{vp}_0$ (no treatment, 32.4\% of states), followed by $\texttt{fl}_4\texttt{vp}_0$ (maximum fluids, no vasopressors, 20.4\%) and $\texttt{fl}_3\texttt{vp}_0$ (14.9\%). Combined high-dose actions ($\texttt{fl}_4\texttt{vp}_4$) are assigned to only 2.8\% of states.

Approximately 21.6\% of patients survive under a purely observational trajectory that never departs from no treatment (Property~6a).
The policy escalates to maximal fluid resuscitation at some point for over half of all patients (Property~6b), while the most aggressive combined intervention is reached in roughly 10\% of trajectories (Property~6c).

\begin{figure}
    \centering
    \scalebox{0.78}{
\begin{tikzpicture}
\begin{axis}[
    xbar,
    width=12cm,
    height=24cm,
    xlabel={Drop in survival probability},
    symbolic y coords={mechvent,Temp\_C,gender,re\_admission,max\_dose\_vaso,input\_total,MeanBP,output\_total,SpO2,HR,Shock\_Index,Creatinine,RR,Potassium,cumulated\_balance,DiaBP,SGOT,Arterial\_pH,Sodium,Hb,Weight\_kg,paCO2,SysBP,output\_4hourly,HCO3,Chloride,WBC\_count,Calcium,Glucose,INR,Platelets\_count,PTT,SGPT,age,BUN,Arterial\_BE,paO2,Magnesium,SOFA,PT,Arterial\_lactate,SIRS,FiO2\_1,Total\_bili,PaO2\_FiO2,GCS,input\_4hourly},
    ytick=data,
    y tick label style={font=\normalsize},
    nodes near coords={\pgfmathprintnumber[fixed, precision=4]{\pgfplotspointmeta}},
    every x tick label/.append style={font=\normalsize},
    label style={font=\large},
    nodes near coords align={horizontal},
    every node near coord/.append style={font=\small, anchor=west},
    xmin=0,
    xmax=0.06,
    bar width=6pt,
    enlarge y limits={abs=0.3cm},
]
\addplot coordinates {
    (0.0503301342,input\_4hourly)
    (0.0060397854,GCS)
    (0.0051973542,PaO2\_FiO2)
    (0.0037256508,Total\_bili)
    (0.0028547250,FiO2\_1)
    (0.0027174530,SIRS)
    (0.0027154261,Arterial\_lactate)
    (0.0026680957,PT)
    (0.0025557715,SOFA)
    (0.0025502975,Magnesium)
    (0.0024536621,paO2)
    (0.0022745864,Arterial\_BE)
    (0.0022404815,BUN)
    (0.0022240972,age)
    (0.0021074902,SGPT)
    (0.0019029541,PTT)
    (0.0018718513,Platelets\_count)
    (0.0016873624,INR)
    (0.0015979387,Glucose)
    (0.0015333438,Calcium)
    (0.0015010746,WBC\_count)
    (0.0014653602,Chloride)
    (0.0014139912,HCO3)
    (0.0013992298,output\_4hourly)
    (0.0013902285,SysBP)
    (0.0012691732,paCO2)
    (0.0012640196,Weight\_kg)
    (0.0012619068,Hb)
    (0.0011546747,Sodium)
    (0.0011117007,Arterial\_pH)
    (0.0010945920,SGOT)
    (0.0010016712,DiaBP)
    (0.0008685052,cumulated\_balance)
    (0.0008245506,Potassium)
    (0.0007814300,RR)
    (0.0007730937,Creatinine)
    (0.0007654157,Shock\_Index)
    (0.0007425100,HR)
    (0.0006920317,SpO2)
    (0.0003917357,output\_total)
    (0.0002681041,MeanBP)
    (0.0001220887,input\_total)
    (0.0000457379,max\_dose\_vaso)
    (0.0000000000,re\_admission)
    (0.0000000000,gender)
    (0.0000000000,Temp\_C)
    (0.0000000000,mechvent)
};
\end{axis}
\end{tikzpicture}
    }
    \caption{Feature importance ranking via feature pruning. Each bar shows the change in survival probability when the corresponding feature $f_i$ is removed from the policy's input. The feature \texttt{input\_4hourly} dominates, indicating the policy relies heavily on the previous dosing window.}
    \label{fig:rankings}
\end{figure}
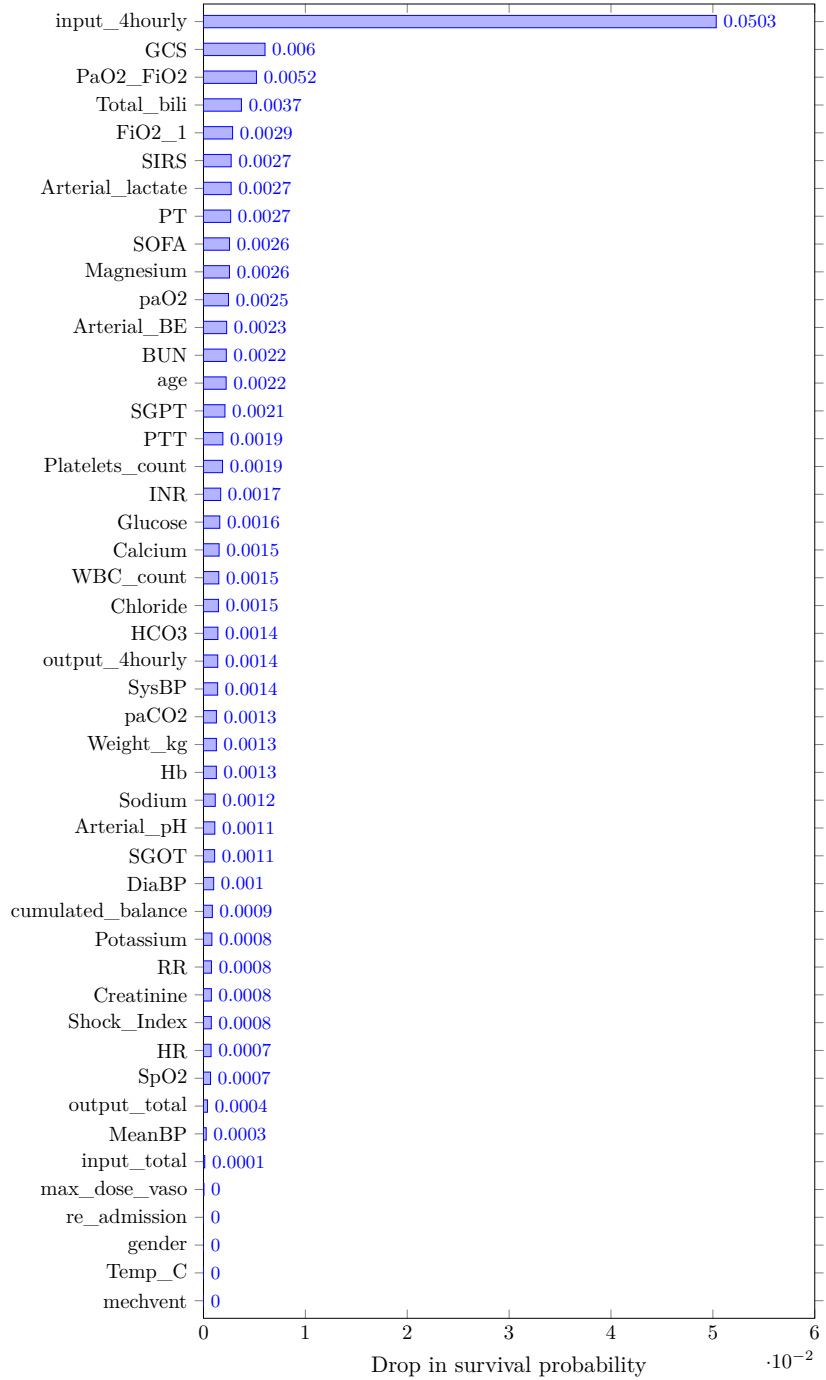

The policy disproportionately relies on \texttt{input\_4hourly}, which encodes the total IV fluid dose from the previous 4-hour window (see Figure~\ref{fig:rankings}).
As a result, the policy may have learned a simple autoregressive dosing heuristic (e.g., continue or adjust based on what was just given) rather than genuinely conditioning treatment decisions on the patient's evolving organ function, which raises concerns about generalization beyond the modeled MDP.
When cross-checking with feature permutation importance, we find that \texttt{input\_4hourly} is ranked as the most important feature in 93.4\% of states, and the probability of reaching a state where it is the most important feature is 1.0 (Property~7a).

\begin{table}[ht]
\centering
\caption{Trained policy (induced DTMC $D^\pi$): baseline outcomes and treatment trajectory properties. Action labels $\texttt{fl}_i\texttt{vp}_j$ encode IV-fluid level $i$ and vasopressor level $j$ ($0$ = none, $4$ = maximum).}
\label{tab:part-b-baseline}
\small
\begin{tabular}{cp{5.2cm}lr}
\toprule
\textbf{ID} & \textbf{Description} & \textbf{PCTL Property} & \textbf{Prob.} \\
\midrule
5a & Survival probability
   & $P_{=?}(\lozenge\;\texttt{survived})$
   & 0.8751 \\
5b & Death probability
   & $P_{=?}(\lozenge\;\texttt{died})$
   & 0.1249 \\
6a & Survive via no-treatment path
   & $P_{=?}(\texttt{fl0\_vp0}\;\mathcal{U}\;\texttt{survived})$
   & 0.2159 \\
6b & Reach max fluids, no vasopressors
   & $P_{=?}(\lozenge\;\texttt{fl4\_vp0})$
   & 0.5414 \\
6c & Reach max fluids + max vasopressors
   & $P_{=?}(\lozenge\;\texttt{fl4\_vp4})$
   & 0.1029 \\
7a & Reach state with \texttt{input\_4hourly} as most important feature
   & $P_{=?}(\lozenge\;\texttt{imp\_input\_4hourly})$
   & 1.0 \\
\bottomrule
\end{tabular}
\end{table}

\emph{Summary.} \emph{COOL-MC} goes beyond standard RL evaluation by formally calculating survival probability and quantitatively characterizing treatment trajectories via PCTL. Crucially, it is the combination of explainability methods with probabilistic model checking that enables precise analysis: feature pruning and feature permutation importance identify which features drive the policy's decisions, while PCTL queries formalize these findings as verifiable properties over the induced DTMC, revealing that the policy relies almost entirely on prior dosing history rather than the patient's evolving clinical state.
Such weaknesses would remain hidden behind the policy's optimal aggregate performance without \emph{COOL-MC}'s integration of formal verification and explainability.

\section{Discussion}
This section discusses the capabilities that \emph{COOL-MC} demonstrates in this case study.

\subsection{The Role of COOL-MC in Full MDP Analysis}
A natural question is why \emph{COOL-MC} is needed at all for the full-MDP analysis in Part~A, given that the ICU-Sepsis MDP is small enough for direct model checking with tools such as PRISM~\cite{KNP11} or Storm~\cite{DBLP:journals/sttt/HenselJKQV22}.
The answer lies in the gap between the PRISM model and the clinical semantics of the states it~encodes.

In our PRISM encoding, each state is represented solely by an integer identifier (see Figure~\ref{fig:state-representation}).
This compact encoding is a practical necessity: encoding all 47 discretized clinical features as explicit PRISM state features/variables, together with their transitions, produces model files of several gigabytes, which cause model checkers to exhaust available memory before the model can even be loaded.
The integer encoding reduces the PRISM file size, substantially lowering memory requirements for both model construction and model checking.

However, this efficiency comes at a cost in expressiveness.
The integer representation is sufficient for basic structural queries, such as computing the maximum or minimum probability of reaching survival or death, because these properties refer only to labels that can be attached directly to the corresponding state identifiers.
For any analysis that involves clinical conditions, such as whether a patient passes through a high SOFA state, the integer encoding alone is insufficient.
Properties like $P_{\max}(\neg\textit{high\_sofa}\ \mathcal{U}\ \textit{survived})$ require that states be annotated with atomic propositions derived from their underlying 47-dimensional feature vectors, which the compact PRISM model does not natively carry.

\emph{COOL-MC} bridges this gap through its automated state labeling mechanism.
During model construction, \emph{COOL-MC} applies the observation function $\mathbb{O}$ to map each integer state identifier to its corresponding observation and evaluates user-specified labeling rules (e.g., label a state as \textit{high\_sofa} if its SOFA value exceeds a given threshold).
The resulting atomic propositions are injected into the built model, making clinically meaningful PCTL properties amenable to model checking without requiring the user to manually inspect and annotate hundreds of states.
This capability enables the analyses reported in Table~\ref{tab:severity_subgroup}.

The same labeling mechanism applies equally to the full MDP and to the policy-induced DTMC analyzed in Part~B, providing a unified framework for both global bounds analysis and policy-specific verification.

\subsection{Hard Bounds as a Calibration Framework}
With feature-based state labels in place, the full-MDP analysis in Part~A provides a calibration framework against which any trained policy in the RL sepsis benchmark~\cite{choudhary2024icu} can be assessed.
The gap between best-case and worst-case probabilities (e.g., 0.8751 vs.\ 0.633 for survival) quantifies the influence of treatment decisions.
Together, they provide reference points for evaluating any RL-based treatment policy on this MDP.

\subsection{From Full MDP to Policy-Specific Analysis}
While the full-MDP analysis establishes theoretical bounds, it does not characterize how a specific trained policy behaves or explain its decision-making.
Moreover, full-MDP model checking will not always be feasible: for larger environments, the state-space explosion problem renders exhaustive analysis intractable.

The core architectural advantage of \emph{COOL-MC} addresses this.
Its analysis pipeline can operate on a policy-induced DTMC without building the full MDP.
Since the induced DTMC contains only the states reachable under a single deterministic policy, it can be smaller than the full MDP, making formal verification feasible even when exhaustive MDP analysis is not.
The ICU-Sepsis case study validates this approach: because both full-MDP and policy-specific analyses are possible on the same benchmark, we can confirm that \emph{COOL-MC}'s policy-level findings are consistent with the global bounds, and the number of transitions is reduced, building confidence in its application to settings where only the policy-level analysis is feasible.

\subsection{From Aggregate Performance to Structural Understanding}
Standard RL evaluation reduces a policy to a single scalar: expected reward.
In the ICU-Sepsis benchmark, this corresponds to reporting a survival probability of 0.8751, matching the theoretical optimum.
This number does not reveal \emph{how} the policy achieves that outcome.
\emph{COOL-MC} moves beyond this aggregate view by constructing the induced DTMC and making it available for PCTL queries, state labeling, and explainability analysis, transforming the evaluation into a structured characterization of the policy's behavior across all reachable treatment trajectories.

For instance, the analysis shows that 21.6\% of patients survive under a purely observational trajectory (Property~6a), that maximal fluid resuscitation is reached in over half of all trajectories (Property~6b), and that the most aggressive combined intervention is applied in roughly 10\% of cases (Property~6c).
These findings are not accessible through standard RL evaluation and can be used to conduct an extended analysis of trained RL policies within the ICU-Sepsis~benchmark.

\subsection{Exposing Hidden Policy Weaknesses}
The trained policy relies predominantly on \texttt{input\_4hourly}, the total IV fluid dose administered in the preceding 4-hour window.
Policy feature pruning with probabilistic model checking shows that removing this single feature causes the largest degradation in survival probability, and feature permutation importance confirms that it is ranked as the most important feature in 93.4\% of states.
This indicates that the policy has learned an autoregressive dosing heuristic: it selects the next action primarily based on the previous dosing rather than on other state features.

This finding raises a generalization concern.
A policy that relies on prior dosing rather than responding to other state features may perform well within the training MDP, where dosing history and clinical state are correlated by construction, but could fail when this correlation does not hold.
This weakness is not detectable through standard evaluation, since the policy achieves the optimal survival probability.
The combination of feature pruning to identify the dominant input features, PCTL queries to quantify its prevalence across trajectories, and the induced DTMC to make these analyses exact exposes this structural property of the policy.

\subsection{COOL-MC as a Pre-Deployment Analysis Tool}
The investigations in Parts~A and~B illustrate how \emph{COOL-MC} supports systematic analysis of sepsis treatment policies before~deployment.

PCTL queries can formalize specific concerns as verifiable properties.
By labeling states with appropriate atomic propositions and writing the corresponding PCTL formula, questions such as ``What is the probability that the policy withholds vasopressors from a patient in a low blood pressure state?'' become formally answerable on the induced DTMC with exact~probabilities.

The combination of state labeling and feature-importance permutation ranking enables trajectory-level analysis of the policy's decision-making: which features drive the decision at each state and how the importance ranking changes as the patient's state evolves along the policy trajectory.

Feature pruning complements this with a global sensitivity analysis that identifies reliance on specific inputs or confirms that the policy integrates information from multiple feature groups.
In this case study, the finding that a single dosing-history feature drives the policy could inform retraining with other loss functions to reduce reliance on that feature.

\subsection{Additional COOL-MC Capabilities}\label{sec:extra}
In this work, we focused on safe RL with shielding, PCTL verification on induced DTMCs, feature pruning, and temporal feature-importance permutation ranking. However, COOL-MC offers additional capabilities that could further enrich the analysis of sepsis treatment policies.

COOL-MC supports counterfactual reasoning~\cite{gross2024enhancing}, where states leading to safety violations are identified through model checking and passed to an expert who explains why the policy's action was problematic and proposes an alternative.

Additionally, COOL-MC's feature abstraction and feature remapping capabilities~\cite{DBLP:conf/setta/GrossJJP22} allow analysis of policy behavior when state features are coarsened or transformed.
In a clinical context, this could simulate scenarios where certain lab values are unavailable or measured at lower precision, yielding bounds on policy performance under degraded observations.

COOL-MC also supports robustness verification through adversarial state-observation perturbations~\cite{DBLP:conf/icaart/GrossS0023,DBLP:conf/aips/GrossS0023}.
By perturbing the observation fed to the policy and re-verifying the induced DTMC, one can assess whether small changes in patient measurements cause the policy to select different treatments. In a clinical setting, this could reveal whether noisy or imprecise sensor readings lead to unsafe treatment decisions, providing a formal measure of the policy's robustness to observation uncertainty.

\section{Conclusion}
We applied \emph{COOL-MC} to the ICU-Sepsis benchmark, combining probabilistic model checking with safe and explainable RL to analyze sepsis treatment policies.
Using the full MDP, we established hard bounds on achievable outcomes of RL policies within the benchmark and showed that treatment decisions can meaningfully influence survival.
We trained a safe RL policy that achieves optimal survival probability and verified its behavior on the induced DTMC, demonstrating how PCTL properties, state labeling, and feature pruning can surface temporal, interpretable information about learned treatment strategies.
While the ICU-Sepsis MDP is intended as an RL benchmark and our results should not guide clinical practice, this work shows how formal verification and explainability methods can complement RL in safety-critical healthcare settings.

\emph{Future work} could explore applying this methodology to larger, more realistic sepsis MDPs, ideally constructed from contemporary patient data, with clinician involvement in defining safety specifications and evaluating the clinical plausibility of the resulting policies.
Additionally, integrating \emph{COOL-MC's} capabilities into existing sepsis early-detection tools, such as \emph{Sepsis Watch}\cite{sendak2020real}, could further enhance their predictive and decision-support capabilities.
Furthermore, the additional COOL-MC capabilities discussed in (\S\ref{sec:extra}) offer promising directions: counterfactual LLM reasoning~\cite{gross2024enhancing} could provide clinician-readable explanations of treatment failures, feature abstraction and remapping~\cite{DBLP:conf/setta/GrossJJP22} could assess policy robustness under coarsened or missing observations, and adversarial perturbation analysis~\cite{DBLP:conf/icaart/GrossS0023,DBLP:conf/aips/GrossS0023} could evaluate sensitivity to noisy measurements.

\bibliographystyle{plain}
\bibliography{refs}

\end{document}